\documentclass{mva_style}
\usepackage{graphicx}
\usepackage{color}
\usepackage{amsmath}
\usepackage{url} 

\finalcopy 

\begin{document}
\title{Semantic Segmentation of iPS Cells: Case Study on Model Complexity in Biomedical Imaging}

\author{
  Maoquan Zhang$^{1,2}$ \\
  Bisser Raytchev$^{1}$ \\
  Xiujuan Sun$^{2}$ \\
  \\
  $^1$\parbox[t]{0.9\textwidth}{Graduate School of Advanced Science and Engineering, Hiroshima University, Hiroshima 739-8511, Japan}\\
  $^2$\parbox[t]{0.9\textwidth}{Department of Computer Science, Weifang University of Science and Technology, Shouguang 262700, China}\\
  \\
  \texttt{zhang-maoquan@hiroshima-u.ac.jp, bisser@hiroshima-u.ac.jp, xjs1981524@wfust.edu.cn}
}

\maketitle

\section*{\centering Abstract}
\textit{
  Medical image segmentation requires not only accuracy but also robustness under challenging imaging conditions. In this study, we show that a carefully configured DeepLabv3 model can achieve high performance in segmenting induced pluripotent stem (iPS) cell colonies, and, under our experimental conditions, outperforms large-scale foundation models such as SAM2 and its medical variant MedSAM2—without structural modifications. These results suggest that, for specialized tasks characterized by subtle, low-contrast boundaries, increased model complexity does not necessarily translate to better performance. Our work revisits the assumption that ever-larger and more generalized architectures are always preferable, and provides evidence that appropriately adapted, simpler models may offer strong accuracy and practical reliability in domain-specific biomedical applications. We also offer an open-source implementation that includes strategies for small datasets and domain-specific encoding, with the aim of supporting further advances in semantic segmentation for regenerative medicine and related fields.
}

\section{Introduction}
Induced pluripotent stem (iPS) cells have enabled significant advances in regenerative medicine by allowing reprogrammed somatic cells to differentiate into nearly any cell type~\cite{powell2023automated}. However, the large-scale cultivation of iPS cells requires reliable identification and isolation of healthy colonies, a task that remains manual and subjective in many settings. Automated segmentation holds promise for objective monitoring of colony growth and quality, but phase-contrast images of iPS colonies often present diffuse boundaries, irregular shapes, and intricate textures~\cite{powell2023automated}. Traditional convolutional neural networks, such as U-Net variants, can perform well on clear images but tend to underperform in regions where boundaries are ambiguous. For example, their Dice scores may decline from approximately 0.84 to below 0.7 as boundary clarity diminishes. Iwamoto et al.~\cite{iwamoto2021} attempted to address these uncertainties using Bayesian deep networks and curriculum learning, reporting improved robustness, but overall segmentation accuracy for iPS colonies remained limited.

\begin{figure}[t]
\centering
  \includegraphics[width=\linewidth]{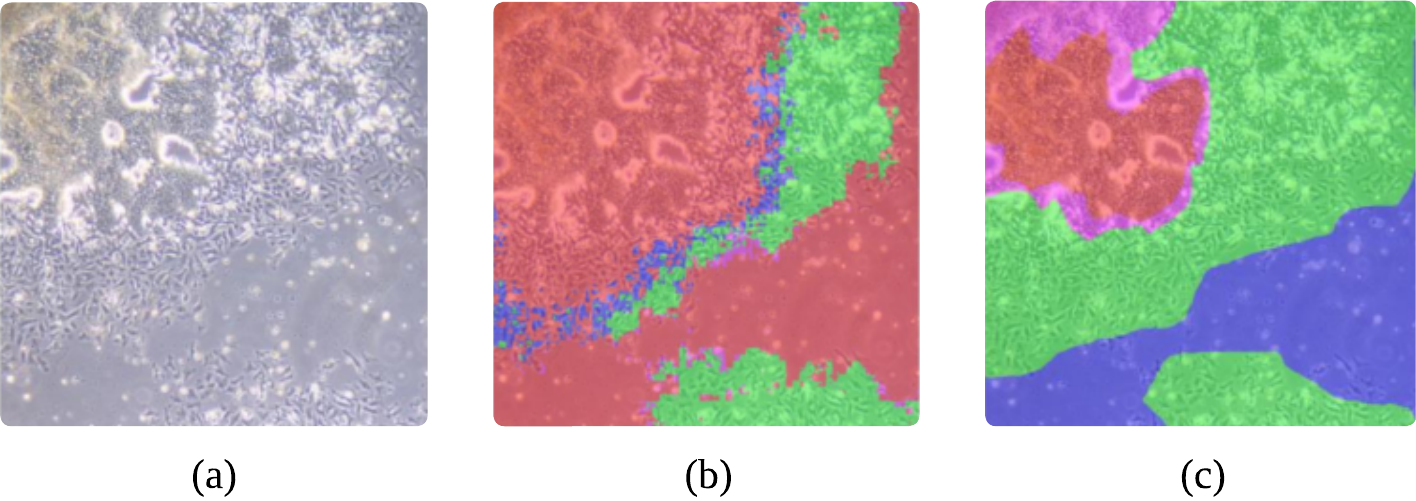}
  \caption{(a) A representative raw iPS image from our test set. (b) A foundation model’s segmentation mask overlaid on the same image. (c) Our specialized DeepLabv3 output, showing markedly improved boundary fidelity in circled regions. In these overlays, red indicates “Good” colonies, green “Bad” colonies, blue culture medium, and pink denotes uncertain areas.}
  \label{f1}
\end{figure}

In parallel, the field has witnessed the emergence of increasingly sophisticated deep models—including TransUNet~\cite{chen2024transunet}, nnU-Net~\cite{isensee2024nnu}, and first-generation foundation models such as SAM and MedSAM~\cite{sam1,ma2024segment}—that have improved overall accuracy. However, these models often produce deterministic outputs and may not fully account for ambiguous or subtle boundaries~\cite{powell2023automated}. Recent models like SAM2~\cite{ravi2024sam2segmentimages} and BioSAM2~\cite{yan2024biomedical} introduce features such as video memory and biomedical-specific fine-tuning (Fig.~\ref{f1}(b)), yet segmentation of iPS colonies remains challenging. As illustrated in Fig.~\ref{f1}(a), extremely faint colony boundaries can still lead to notable segmentation errors even with the latest architectures.

To address these challenges, we employ a DeepLabv3 network~\cite{chen201deeplab3} carefully adapted for iPS imagery. As shown in Fig.~\ref{f1}(c), this approach outperforms both established CNNs (TransUNet, nnU-Net) and recent foundation models (SAM2, BioSAM2), suggesting that a well-adapted, task-focused model can match or exceed more complex architectures for specialized biomedical segmentation. Our main contributions are as follows:
\begin{itemize}
    \item \textbf{Consistent and Accurate iPS Colony Segmentation:} Our DeepLabv3 model achieves high accuracy at the pixel level for iPS cell colonies across all tested settings, addressing the unique challenges of phase-contrast iPS imagery.
    \item \textbf{Re-examining Model Complexity:} Our results indicate that, for this task, a refined, smaller-scale CNN can outperform larger foundation models, highlighting the importance of model adaptation over scale alone.
    \item \textbf{Model Analysis:} We provide insights into why architectural features such as atrous convolutions and receptive field expansion benefit fine-grained cell delineation in this context.
    \item \textbf{Open-Source Toolkit:} We make all code and trained models publicly available, including dataset handling and strategies for working with small sample sizes, to support further research and practical use~\cite{ips_segmentation_github}.
\end{itemize}

\section{Model Comparisons and Analysis}
Building on the objectives outlined in the Introduction, this section presents a comparative evaluation of SAM2, MedSAM2, DeepLabv3, and other reference architectures, including TransUNet, nnU-Net, and BioSAM2. The experiments were designed to examine the trade-offs between generic foundation models and specialized, task-adapted networks. Comparative analysis highlights structural and computational factors that affect segmentation performance and provides context for understanding DeepLabv3’s effectiveness in phase-contrast iPS cell segmentation.

\subsection{Experimental Setup and Key Comparisons} We performed a series of controlled experiments to evaluate each model’s performance and computational efficiency on the iPS colony segmentation task. All experiments used a standardized framework to ensure fair comparison: identical training dataset size, optimizer configuration, input normalization, and evaluation metrics~\cite{ma2024segment,ravi2024sam2segmentimages,chen201deeplab3}.

Experiments were conducted with a fixed random seed (42), using a single \emph{NVIDIA GeForce RTX 3090} GPU and an \emph{Intel Xeon W-2223 CPU @ 3.60~GHz}. For comparison, we report the following metrics for SAM2, MedSAM2, and DeepLabv3 (DLV3): \textit{BatchSize} (mini-batch size per iteration), \textit{T(s)/Epoch} (average seconds per training epoch), \textit{GPU(MiB)} (peak GPU memory usage during training), \textit{Epochs} (number of epochs to reach peak accuracy), and \textit{Acc(IOU)} (Intersection over Union for final segmentation performance).

\begin{table}[t]
  \caption{Training metrics for SAM2, MedSAM2, and DeepLabv3 on the iPS dataset (mean $\pm$ std over five runs). All models were evaluated under identical conditions.}
  \begin{center}
    \begin{tabular}{c | c c c}
      \hline
      \hline
      \makebox[12mm]{Metric} & \makebox[12mm]{SAM2} & 
      \makebox[12mm]{MedSAM2} & \makebox[12mm]{DLV3}\\
      \hline
      BatchSize & 2 & 2   & 2 \\
      T(s)/Epoch & 600 & 581 & 60 \\
      GPU(MiB) & 2,926 & 15,149 & 1,467 \\
      Epochs & 1,000 & 1,000 & 50 \\
      Acc(IOU) & 81.0 $\pm$ 1.56 & 63.5 $\pm$ 2.17 & 97.5 $\pm$ 2.21 \\
      \hline
      \hline
    \end{tabular}
    \label{tab:comparative-params}
  \end{center}
\end{table}

Table~\ref{tab:comparative-params} summarizes training efficiency and resource usage for each model. All models were run using the same batch size and hardware configuration. DeepLabv3 converges in 50 epochs and achieves consistently high IoU scores ($97.5 \pm 2.21$\%), while SAM2 and MedSAM2 require 1,000 epochs and considerably more GPU memory (up to 15~GB for MedSAM2), yet yield lower IoU ($81.0 \pm 1.56$\% and $63.5 \pm 2.17$\%, respectively). These results, averaged over five independent runs, demonstrate both the computational efficiency and robustness of the task-adapted convolutional approach. As shown in Fig.~2, DeepLabv3 predictions closely match ground truth labels, effectively segmenting subtle colony boundaries even in challenging cases. While SAM2 and MedSAM2 demand greater computational resources and extended training, their final segmentation accuracies remain limited in this context.

\begin{figure}[t]
\centering
  \includegraphics[width=\linewidth]{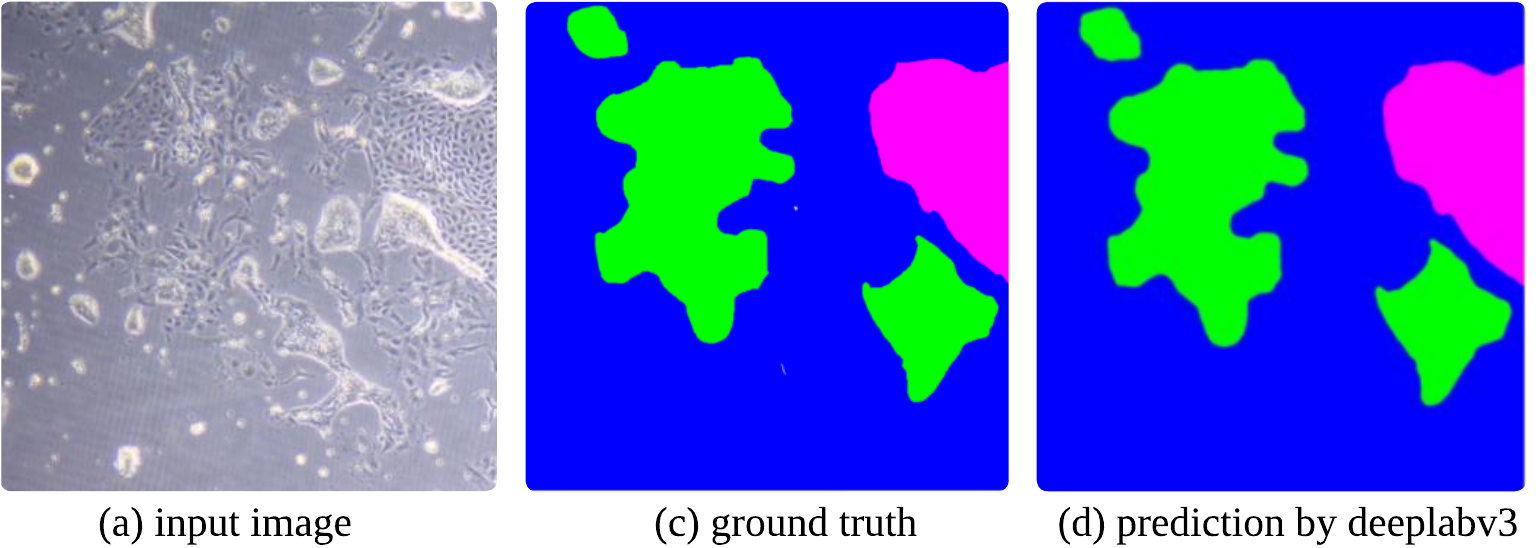}
  \caption{Visual comparison of (a) input iPS image, (b) ground truth segmentation, and (c) DeepLabv3 prediction. Green indicates ``Good'' colonies, pink marks uncertain regions, and blue denotes background. The prediction (c) shows close alignment with the ground truth (b), including regions with subtle or ambiguous boundaries.}
  \label{f2}
\end{figure}

\subsection{Dataset Setup and Analysis}
A total of 60 high-resolution iPS images were used in this study. Images were divided into patches of size $512\times512$ and $1024\times1024$, and then split into training, validation, and test sets in a 6:3:1 ratio at the image level. For \textbf{SAM2}, both patch sizes produced comparable IoU values; for \textbf{MedSAM2}, only $1024\times1024$ patches yielded stable training. \textbf{DeepLabv3} demonstrated robust performance across both patch sizes, converging within 50 epochs. For consistency, all results in Table~\ref{tab:comparative-params} are reported using a standardized subset of 590 patch images of size $1024\times1024$.

As shown in Fig.~\ref{f2}, DeepLabv3 segmentation closely matches the ground truth, even in regions with faint boundaries and minor annotation artifacts. This visual alignment further supports the model’s robustness and accuracy for iPS colony segmentation under challenging imaging conditions.

\subsection{Insights from Architecture and Resource Analysis}
To further interpret the comparative results, we examine the structural factors underlying the observed performance differences between DeepLabv3 and the SAM2/MedSAM2 models. Our analysis indicates that DeepLabv3’s design—featuring atrous convolutions and multi-scale feature aggregation—is particularly effective for static phase-contrast iPS cell images. In contrast, SAM2’s memory-based video segmentation architecture introduces additional computational overhead without notable advantage in this setting.

\subsubsection{DeepLabv3: Efficiency with ASPP and Dilated Convolutions}
\begin{figure}[t]
\centering
  \includegraphics[width=\linewidth]{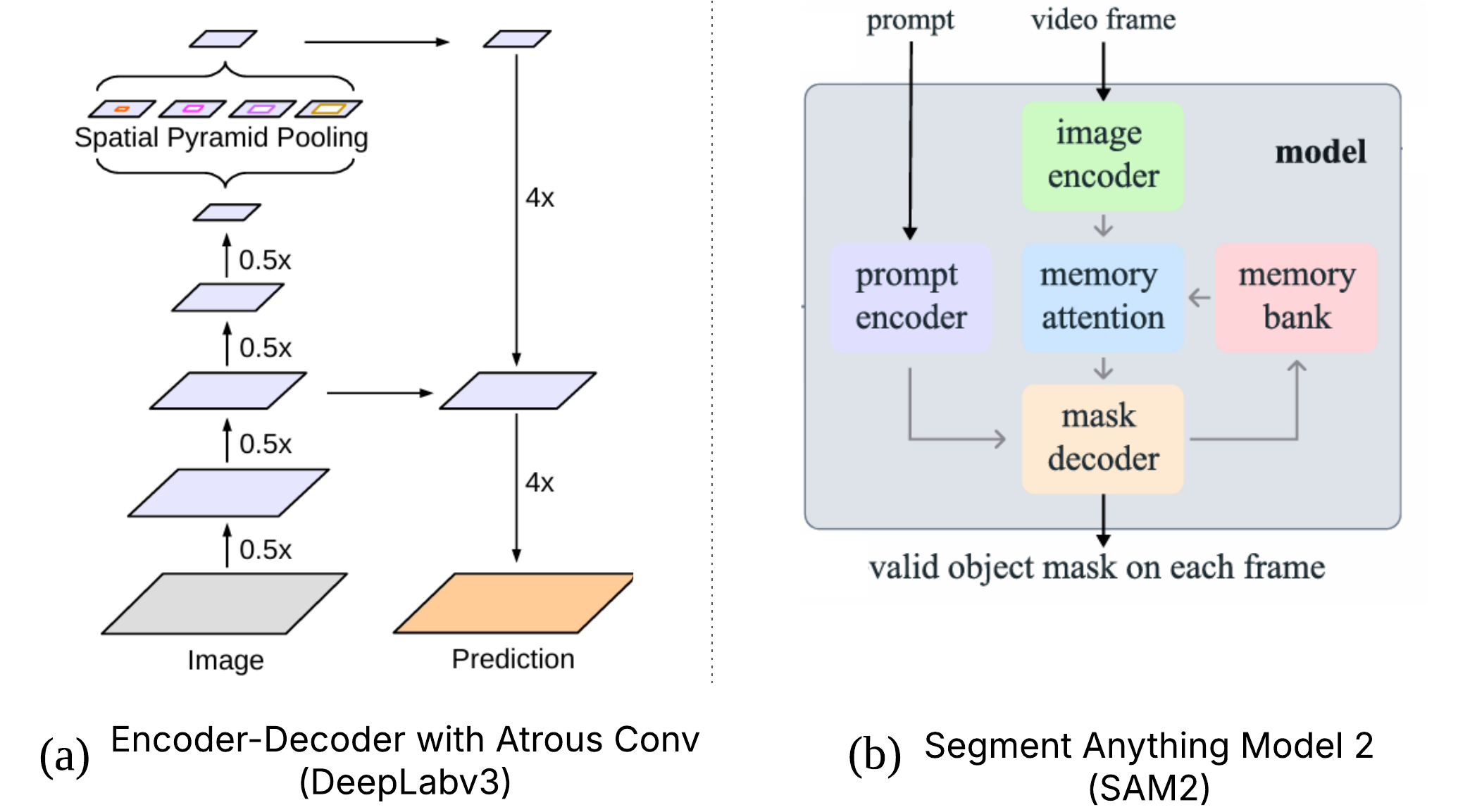}
  \caption{Core modules of (a) DeepLabv3~\cite{chen201deeplab3} and (b) SAM2~\cite{ravi2024sam2segmentimages}. DeepLabv3 leverages spatial pyramid pooling and atrous convolutions for efficient feature extraction, while SAM2 incorporates a memory bank and video-specific modules optimized for temporal segmentation.}
  \label{f3}
\end{figure}

A central component of DeepLabv3’s performance is the Atrous Spatial Pyramid Pooling (ASPP) module (Fig.~\ref{f3}(a)), which combines multiple parallel dilated (atrous) convolutions with varying rates~\cite{chen201deeplab3}. These dilated kernels expand the receptive field, enabling the network to merge global context (such as overall colony shape) and local features (such as fine edge gradients) without increasing parameter count~\cite{chen2017deeplab}. This multi-scale approach is especially advantageous for the low-contrast and ambiguous boundaries present in iPS colony images, as illustrated in Fig.~\ref{f1}(c).

Another notable advantage is the moderate parameter count in DeepLabv3’s backbone (e.g., ResNet-50 with approximately 42 million parameters~\cite{trainWithDeepLabv3}). The use of atrous convolutions allows DeepLabv3 to maintain high-resolution feature maps and efficient training, without the need for deeper or more complex networks. In our experiments, DeepLabv3 achieved consistently high IoU scores within 50 epochs and with limited GPU resources (see Table~\ref{tab:comparative-params}). In contrast, SAM2 required over 1,000 epochs and substantially higher GPU memory to reach its best accuracy (about 82\%) under the same settings. These results suggest that, for specialized biomedical segmentation tasks, a carefully tuned, moderate-sized network can be both effective and efficient compared to larger, general-purpose models.
\subsubsection{SAM2: Memory-Based Mechanisms for Streaming Video}
SAM2 (“Segment Anything Model 2”; Fig.~\ref{f3}(b)) extends the original architecture to video tasks by incorporating a memory module~\cite{ravi2024sam2segmentimages, ma2024segment} on top of a Transformer backbone with several hundred million parameters~\cite{sam1}. While this approach is well suited for multi-frame applications that require temporal tracking, it introduces substantial computational overhead for single-frame segmentation tasks such as static iPS colony images, resulting in increased GPU usage and longer training times. In our experiments, SAM2 required more than 1,000 epochs to reach moderate accuracy (see Table~\ref{tab:comparative-params}), indicating a mismatch between memory-centric architectures and the requirements of static biomedical imaging.

A further limitation is SAM2’s pretraining on large-scale, high-contrast video datasets~\cite{oh2019video}, which differs significantly from the low-contrast, fine-grained boundaries in iPS images. The memory bank and prompt-based segmentation~\cite{sam1} offer limited benefit in this context, where historical or temporal information is minimal. Consequently, even with extensive fine-tuning, SAM2 does not match the segmentation accuracy of more specialized, streamlined models such as DeepLabv3.

\subsubsection{Implications for Domain-Specific Medical Imaging}
Foundation models can address diverse tasks, but often require significant adaptation and computational resources to match the accuracy of domain-specific networks~\cite{wang2024dlv3seg, isensee2024nnu}. For iPS colony segmentation, high accuracy relies on resolving small, low-contrast boundaries, where DeepLabv3’s multi-rate dilated convolutions and ASPP are particularly effective. In contrast, SAM2’s memory modules and large Transformer backbone, while suited for video tasks, add complexity without clear benefit for single-frame biomedical images~\cite{ma2024segment, ravi2024sam2segmentimages, zhang2023segmentmodelsammedical}.

MedSAM2~\cite{ma2024segment}, fine-tuned for medical use, performs worse than SAM2 on iPS images, likely due to overfitting to modalities such as CT or MRI~\cite{li2024adapting}. These results highlight the importance of closely matching the target data and fine-tuning set when adapting models for biomedical imaging, as domain-specific adjustments may otherwise reduce performance.

\subsection{Analysis of iPS Cell Segmentation Efficiency}
\subsubsection{Atrous Convolutions in DeepLabv3}
DeepLabv3 employs atrous (dilated) convolutions to expand the receptive field without increasing the number of parameters~\cite{chen2017deeplab}. For a kernel of size $k$ and dilation rate $r$, the effective receptive field becomes $(k-1)r+1$. For example, a $3\times3$ filter with $r=2$ covers a $5\times5$ input region, capturing broader context without additional weights. Stacking dilated layers further increases the field of view, which is essential for accurately segmenting faint and ambiguous iPS cell boundaries.

The atrous spatial pyramid pooling (ASPP) module in DeepLabv3 applies multiple dilation rates in parallel, enabling effective multi-scale feature extraction~\cite{deeplabv3_medical}. Because dilation “stretches” filters by inserting zeros, no new parameters are added~\cite{chen2017deeplab}. This design maintains the efficiency of a standard CNN while providing a large receptive field, supporting both global context and fine detail—key for low-contrast iPS cell images—without overfitting or excessive computation.

\subsubsection{Computational Complexity and Performance of Self-Attention in SAM2}
SAM2’s multi-head self-attention mechanism requires $\mathcal{O}(dN^2)$ operations for $N$ tokens and feature dimension $d$~\cite{keles2023computational}, compared to the $\mathcal{O}(N)$ complexity of typical $f\times f$ convolutions. Despite this higher computational cost, SAM2 (Table~\ref{tab:comparative-params})—even without task-specific fine-tuning—outperforms previous results on iPS segmentation (e.g., $IoU = 0.797$~\cite{iwamoto2021}). However, dense self-attention and large memory structures may be less effective for relatively uniform, low-contrast images such as iPS colonies.

MedSAM2, designed as a medical imaging adaptation of SAM2, exhibits a notable drop in accuracy on iPS cell data. Its complex memory bank and specialized components can introduce unnecessary complexity, highlighting that large, specialized Transformers are not always optimal for all biomedical segmentation tasks.

\subsubsection{Conclusion and Implications}

Our analysis provides three main insights:
\begin{enumerate}
    \item \textbf{DeepLabv3’s Efficiency and Accuracy:} Atrous convolutions allow DeepLabv3 to expand the receptive field to $(k-1)r + 1$ per layer~\cite{chen201deeplab3}, enabling efficient capture of both subtle edges and global context. This contributes to consistently high IoU values (mean $97.5 \pm 2.21$\%) for iPS cell segmentation.

    \item \textbf{Computational Overhead of SAM2’s Self-Attention:} Although SAM2 surpasses previous methods (e.g., IoU $= 0.797$~\cite{iwamoto2021}), its self-attention mechanism (\(\mathcal{O}(dN^2)\),~\cite{keles2023computational}) results in considerable computational cost, which is less suitable for single-frame, homogeneous biomedical data.

    \item \textbf{Domain Misalignment in MedSAM2:} MedSAM2’s memory module and fine-tuning on medical datasets unrelated to iPS cells can lead to feature-space misalignment and reduced performance in this application~\cite{choudhary2020advancing}.
\end{enumerate}

These results highlight the benefits of compact, well-adapted models such as DeepLabv3 for specialized biomedical segmentation tasks. Future research may focus on lightweight transformer architectures and more targeted domain adaptation to further improve segmentation efficiency and robustness in complex medical imaging scenarios.

\section{Conclusion}
We have shown that DeepLabv3, when properly configured, achieves consistently high accuracy and greater computational efficiency than large-scale foundation models such as SAM2 and MedSAM2 for iPS colony segmentation. Our results highlight the advantages of domain-adapted models, as smaller and targeted networks can more effectively handle faint boundaries and intra-class variability in low-contrast biomedical images. By providing our code and models, we hope to support further research on hybrid and efficient architectures that balance global context and resource requirements. Overall, this study suggests that model design tailored to the specific demands of medical imaging can offer better performance than increasing model scale alone.

\section*{Acknowledgment}
This research was supported by a JSPS KAKENHI Grant Number JP23K11170 to BR.

\bibliographystyle{IEEEtran}
\bibliography{mycite}

\begin{thebibliography}{99}
\bibitem{powell2023automated}
K.~A. Powell, L.~R. Bohrer, N.~E. Stone, B.~Hittle, K.~R. Anfinson, V.~Luangphakdy, G.~Muschler, R.~F. Mullins, E.~M. Stone, and B.~A. Tucker, ``Automated human induced pluripotent stem cell colony segmentation for use in cell culture automation applications,'' \emph{SLAS technology}, vol.~28, no.~6, pp. 416--422, 2023.

\bibitem{chen2024transunet}
J.~Chen, J.~Mei, X.~Li, Y.~Lu, Q.~Yu, Q.~Wei, X.~Luo, Y.~Xie, E.~Adeli, Y.~Wang, \emph{et~al.}, ``TransUNet: Rethinking the U-Net architecture design for medical image segmentation through the lens of transformers,'' \emph{Medical Image Analysis}, vol.~97, p. 103280, 2024.

\bibitem{isensee2024nnu}
F.~Isensee, T.~Wald, C.~Ulrich, M.~Baumgartner, S.~Roy, K.~Maier-Hein, and P.~F. Jaeger, ``nnu-net revisited: A call for rigorous validation in 3d medical image segmentation,'' in \emph{International Conference on Medical Image Computing and Computer-Assisted Intervention}, 2024, pp. 488--498.

\bibitem{isensee2021nnu}
F.~Isensee, P.~F. Jaeger, S.~A.~A. Kohl, J.~Petersen, and K.~H. Maier-Hein, ``nnU-Net: a self-configuring method for deep learning-based biomedical image segmentation,'' \emph{Nature methods}, vol.~18, no.~2, pp. 203--211, 2021.

\bibitem{iwamoto2021}
S.~Iwamoto, B.~Raytchev, T.~Tamaki, and K.~Kaneda, ``Improving the reliability of semantic segmentation of medical images by uncertainty modeling with Bayesian deep networks and curriculum learning,'' in \emph{Uncertainty for Safe Utilization of Machine Learning in Medical Imaging, and Perinatal Imaging, Placental and Preterm Image Analysis: 3rd International Workshop, UNSURE 2021, and 6th International Workshop, PIPPI 2021, Held in Conjunction with MICCAI 2021, Strasbourg, France, October 1, 2021, Proceedings 3}, 2021, pp. 34--43.

\bibitem{sam1}
A.~Kirillov, E.~Mintun, N.~Ravi, H.~Mao, C.~Rolland, L.~Gustafson, T.~Xiao, S.~Whitehead, A.~C. Berg, W.-Y. Lo, \emph{et~al.}, ``Segment anything,'' in \emph{Proceedings of the IEEE/CVF international conference on computer vision}, 2023, pp. 4015--4026.

\bibitem{ravi2024sam2segmentimages}
N.~Ravi, V.~Gabeur, Y.-T. Hu, R.~Hu, C.~Ryali, T.~Ma, H.~Khedr, R.~Rädle, C.~Rolland, L.~Gustafson, E.~Mintun, J.~Pan, K.~V.~A. Alwala, N.~Carion, C.-Y. Wu, R.~Girshick, P.~Dollár, and C.~Feichtenhofer, ``SAM 2: Segment Anything in Images and Videos,'' 2024. [Online]. Available: \url{https://arxiv.org/abs/2408.00714}

\bibitem{ma2024segment}
J.~Ma, Y.~He, F.~Li, L.~Han, C.~You, and B.~Wang, ``Segment anything in medical images,'' \emph{Nature Communications}, vol.~15, no.~1, p. 654, 2024.

\bibitem{chen201deeplab3}
L.-C. Chen, Y.~Zhu, G.~Papandreou, F.~Schroff, and H.~Adam, ``Encoder-Decoder with Atrous Separable Convolution for Semantic Image Segmentation,'' in \emph{Proceedings of the European Conference on Computer Vision (ECCV)}, September 2018.

\bibitem{yan2024biomedical}
Z.~Yan, W.~Sun, R.~Zhou, Z.~Yuan, K.~Zhang, Y.~Li, T.~Liu, Q.~Li, X.~Li, L.~He, \emph{et~al.}, ``Biomedical sam 2: Segment anything in biomedical images and videos,'' \emph{arXiv preprint arXiv:2408.03286}, 2024.

\bibitem{chen2017deeplab}
L.-C. Chen, G.~Papandreou, I.~Kokkinos, K.~Murphy, and A.~L. Yuille, ``Deeplab: Semantic image segmentation with deep convolutional nets, atrous convolution, and fully connected crfs,'' \emph{IEEE transactions on pattern analysis and machine intelligence}, vol.~40, no.~4, pp. 834--848, 2017.

\bibitem{trainWithDeepLabv3}
V.~Blagojevic, ``Train Neural Net for Semantic Segmentation with PyTorch in 50 Lines of Code,'' 2021. [Online]. Available: \url{https://towardsdatascience.com/train-neural-net-for-semantic-segmentation-with-pytorch-in-50-lines-of-code-830c71a6544f/}

\bibitem{oh2019video}
S.~W. Oh, J.-Y. Lee, N.~Xu, and S.~J. Kim, ``Video object segmentation using space-time memory networks,'' in \emph{Proceedings of the IEEE/CVF international conference on computer vision}, 2019, pp. 9226--9235.

\bibitem{wang2024dlv3seg}
D.~Wang, Z.~Liu, and Y.~Zhang, ``Segmentation of Carotid Plaques Based on Improved DeepLabV3,'' in \emph{Proceeding of the 2024 5th International Conference on Computer Science and Management Technology}, 2024, pp. 201--207.

\bibitem{zhang2023segmentmodelsammedical}
Y.~Zhang and R.~Jiao, ``Towards Segment Anything Model (SAM) for Medical Image Segmentation: A Survey,'' 2023. [Online]. Available: \url{https://arxiv.org/abs/2305.03678}

\bibitem{li2024adapting}
K.~Li and P.~Rajpurkar, ``Adapting segment anything models to medical imaging via fine-tuning without domain pretraining,'' in \emph{AAAI 2024 Spring Symposium on Clinical Foundation Models}, 2024.

\bibitem{deeplabv3_medical}
IMAIOS, ``DeepLabV3 and Medical Imaging,'' 2022. [Online]. Available: \url{https://www.imaios.com/en/resources/blog/deeplabv3-and-medical-imaging}

\bibitem{keles2023computational}
F.~D. Keles, P.~M. Wijewardena, and C.~Hegde, ``On the computational complexity of self-attention,'' in \emph{International conference on algorithmic learning theory}, 2023, pp. 597--619.

\bibitem{domain_adaptation_v7labs}
V7 Labs, ``Domain Adaptation Guide,'' 2023. [Online]. Available: \url{https://www.v7labs.com/blog/domain-adaptation-guide}

\bibitem{wang2024samcl}
Z.~Wang, K.~Ji, D.~Wang, and F.~Cheng, ``SAMCL: Empowering SAM to Continually Learn from Dynamic Domains,'' \emph{arXiv preprint arXiv:2412.05012}, 2024.

\bibitem{choudhary2020advancing}
A.~Choudhary, L.~Tong, Y.~Zhu, and M.~D. Wang, ``Advancing medical imaging informatics by deep learning-based domain adaptation,'' \emph{Yearbook of medical informatics}, vol.~29, no.~01, pp. 129--138, 2020.

\bibitem{ips_segmentation_github}
Z.~Maoquan, ``iPS-Semantic-Segmentation,'' 2025. [Online]. Available: \url{https://github.com/afountain/iPS-Semantic-Segmentation}
\end{thebibliography}

\end{document}